\documentclass[10pt,twocolumn,letterpaper]{article}

\usepackage[pagenumbers]{cvpr} 

\usepackage{graphicx}
\usepackage{amsmath}
\usepackage{amssymb}
\usepackage{booktabs}

\usepackage[normalem]{ulem}
\usepackage{ifthen}
\usepackage{enumitem}
\usepackage{multicol}
\usepackage{multirow}
\newcommand{\anna}[2][]{%
    \ifthenelse{ \equal{#1}{} }
        {\textcolor{blue}{(Anna) #2}}
        {\textcolor{blue}{(Anna) \sout{#1\xspace}#2}}
}
\newcommand{\tali}[2][]{%
    \ifthenelse{ \equal{#1}{} }
        {\textcolor{magenta}{(Tali) #2}}
        {\textcolor{magenta}{(Tali) \sout{#1\xspace}#2}}
}
\newcommand{\shir}[2][]{%
    \ifthenelse{ \equal{#1}{} }
        {\textcolor{cyan}{(Shir) #2}}
        {\textcolor{cyan}{(Shir) \sout{#1\xspace}#2}}
}
\newcommand{\todo}[2][]{%
    \ifthenelse{ \equal{#1}{} }
        {\textcolor{red}{TODO: #2}}
        {\textcolor{red}{TODO: \sout{#1\xspace}#2}}
}

\usepackage[pagebackref,breaklinks,colorlinks]{hyperref}

\usepackage[capitalize]{cleveref}
\crefname{section}{Sec.}{Secs.}
\Crefname{section}{Section}{Sections}
\Crefname{table}{Table}{Tables}
\crefname{table}{Tab.}{Tabs.}

\def\cvprPaperID{*****} 
\def\confName{CVPR}
\def\confYear{2022}

\begin{document}

\title{\vspace{-.2in} Deep Learning Models for Automated Classification of Dog Emotional States from Facial Expressions \vspace*{-5mm}}

\author{Tali Boneh-Shitrit$^1$ \qquad Shir Amir$^2$ \qquad Annika Bremhorst$^{3,4,5}$ \qquad Daniel S. Mills$^4$\\ \qquad Stefanie Riemer$^3$ \qquad Dror Fried$^1$ \qquad Anna Zamansky$^6$\\
\small 
$^1$Open University, Israel \quad
\small
$^2$Weizmann Institute of Science, Israel \quad
\small
$^3$University of Bern, Switzerland \quad
\small
$^4$University of Lincoln, UK \qquad\\
\small
$^5$dogs and science - Institute for Canine Science and Applied Cynology, Germany \qquad
\small
$^6$University of Haifa, Israel \qquad
}
\maketitle

\begin{abstract}
Similarly to humans, facial expressions in animals are closely linked with emotional states. However, in contrast to the human domain, automated recognition 
of emotional states from facial expressions in animals is underexplored, mainly due to difficulties in data collection and establishment of ground truth concerning emotional states of non-verbal users. 
We apply recent deep learning techniques to classify (positive) anticipation and (negative) frustration of dogs on a dataset collected in a controlled experimental setting.
We explore the suitability of different backbones (e.g. ResNet, ViT) under different supervisions to this task, and find that features of a self-supervised pretrained ViT (DINO-ViT) are superior to the other alternatives.
To the best of our knowledge, this work is the first to address the task of automatic classification of canine emotions on data acquired in a controlled experiment. 
\end{abstract}

\section{Introduction}
\label{sec:intro}

It is widely accepted nowadays that animals are  able to experience emotional states \cite{de2016we}. Facial expressions are produced by most mammalian species \cite{diogo2009origin}, and also are assumed to convey information about emotional states \cite{descovich2017facial,diogo2008fish}. Therefore, they are receiving increasing attention as indicators of emotional states in animals, as well as in research on animal emotions \cite{descovich2017facial}. 

In human emotion research the golden standard for objective measurement of facial expressions is the \emph{Facial Action Coding System} (FACS \cite{Ekman1978}). It is an anatomy-based {manual} annotation system which allows describing facial appearance changes based on movements of the underlying facial muscles. With the recent adaptation of FACS to different non-human species, including dogs (DogFACS \cite{waller2013dogfacs}), FACS-related methods have been applied in several studies for the investigation of dog emotional states. 

Several works have used DogFACS to address two emotional states in canines of different valence: positive anticipation and frustration \cite{caeiro2017development,anikka2019,bremhorst2021evaluating}. Caeiro et al.\cite{caeiro2017development} investigated dog facial expressions associated with positive anticipation and frustration in naturalistic settings using online videos, showing that dogs displayed distinctive facial actions in these two states. 
In contrast, Bremhorst et al.\cite{anikka2019} investigated dogs’ facial expressions in the emotional states of frustration and positive anticipation in a controlled experimental setting, standardizing also the dog breed (Labrador Retriever) to reduce the potential effects of morphological variation and extremes on the dogs’ facial expressions. The authors found that the some action units were more common in the positive condition, and some others in the negative one. 
In a follow-up study, Bremhorst et al.\cite{bremhorst2021evaluating} used a similar set-up with new participants, to induce positive anticipation and frustration in two reward contexts: food and toys. The previous results were replicated, and additional facial actions were also found more common in the negative condition. However, none of the identified units could serve alone as potential emotion indicators, providing consistent correct classifications of the associated emotion. 

A major downside for using FACS (and AnimalFACS) for facial expression analysis is that it requires \emph{extensive human annotation and certification}. It is also time consuming, and may be prone to human error or bias\cite{hamm2011automated}. Therefore, automated approaches are an attractive alternative, considered to have even greater objectivity and reliability than manual approaches\cite{Bartlett1999-ta,Cohn2005-mw}.

Indeed, in the human domain automated facial analysis and effective computing are vibrant fields of research. Numerous commercial software tools for automated facial analysis are available, such as FaceReader by Noldus\cite{Lewinski2014-xy}, Affdex\cite{Stockli2018-dt}, EmoVu\cite{Arnold_undated-dj}, and more.  Some researchers consider automated tools to have greater objectivity and reliability than manual coding, eliminating subjectivity and bias\cite{Bartlett1999-ta,Cohn2005-mw}.

In contrast, the animal domain remains under-explored in the context of {automated} emotion recognition. Some reasons for this are outlined in Hummel et al. \cite{hummel2020automatic}: less data is available, and also animal faces vary more in terms of color, shape and texture. Furthermore, due to lack of verbal basis for establishing ground truth, data acquisition and annotation are much more complicated. {To the best of our knowledge, our work is the first to investigate automatic deep-learning methods for canine emotion recognition on a \emph{rigorously created and controlled dataset}.}

Automated facial analysis has so far been addressed in a few species. Pain recognition from facial expressions has been investigated for rodents \cite{sotocina2011rat,tuttle2018deep,andresen2020towards}, sheep\cite{mahmoud2018estimation}, equines \cite{hummel2020automatic,broome2019dynamics,lencioni2021pain}, and cats\cite{feighel}. Action unit recognition was automated for several types of non-human primates \cite{morozov2021automatic,blumrosen2017towards}. In the context of dogs, Ferres et al. studied automated pose estimation using DeepLabCut \cite{mathis2018deeplabcut} for classification of emotional states as anger, fear, happiness and relaxation \cite{ferres2022predicting}. Franzoni et al.~\cite{franzoni2019preliminary} used a pre-trained CNN (i.e. AlexNet \cite{alexnet}) to classify dog emotional states of joy and anger. However, in both of these works the datasets included images collected from  the internet and annotated by the authors, \emph{with a significant possibility of human bias and inaccurate annotations}. In contrast, our work is based on a dataset obtained in a \emph{controlled setting}, in which positive anticipation and frustration are induced using carefully designed experimental protocol specified in \cite{bremhorst2019differences}.

Liu et al. addressed automated classification of dog breed using part localization \cite{liu2012dog}, highlighting the great challenge in computer vision of dealing with wide variation in shape and appearance in different dog breeds. The dataset used in this paper relaxes this problem by standardizing the breed of the participants to Labrador Retriever.

In this work we investigate the adequacy of \emph{different pre-trained models} for the delicate task of emotion recognition from facial images. We explore the use of two commonly used architectures: ResNet~\cite{he16resnet}, a CNN with residual connections; and Vision Transformer (ViT)~\cite{dosovitskiy2020vit}, a model based on the attention mechanism. We further consider the effect of \emph{different pre-training techniques} on ImageNet~\cite{deng09imagenet}, comparing \emph{supervised} training for image classification with \emph{self-supervised} training using DINO~\cite{caron2021emerging}. Amir et al.~\cite{amir2021deep} show DINO-ViT features encode powerful semantic information on object parts, and animal parts in particular; which can be beneficial to address subtle facial movement analysis for canine emotion classification. To the best of our knowledge, we are the first to explore different backbones for this task.

Our main contributions are the following:
\begin{itemize}
    \item We automatically classify dog emotional states of positive anticipation and frustration based on facial images obtained in a \emph{controlled experimental setting}, without using DogFACS annotations.
    \item We explore the suitability of different pre-trained backbones for this task.
    \item We conclude that DINO-ViT features are most suitable for this classification task, and strengthen this claim using qualitative interpretability methods.
\end{itemize}

\begin{figure}[t!]
\centering
\includegraphics[width=0.5\textwidth]{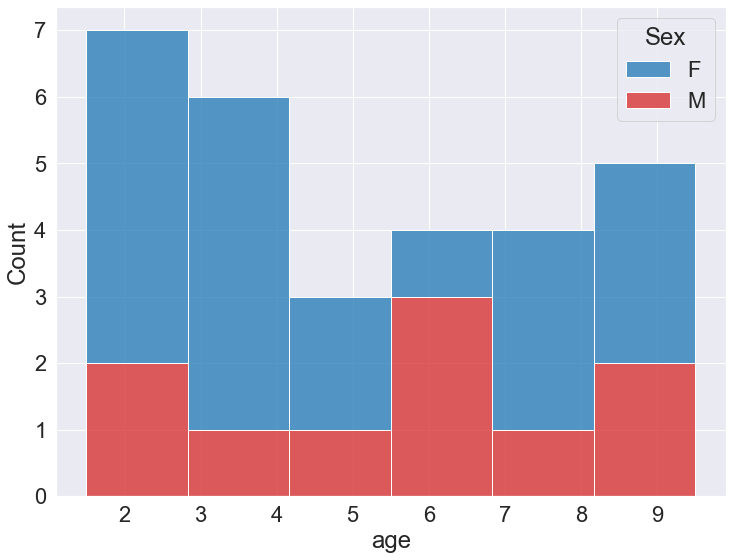}
\caption{\textbf{Number of dogs by age and by sex.} The dataset contains slightly more female than male dogs, and slightly more younger dogs than older ones.}\label{age_dog_sex}\vspace{-0.25em}
\end{figure}

\begin{figure}[t!]
\centering
\includegraphics[width=0.475\textwidth]{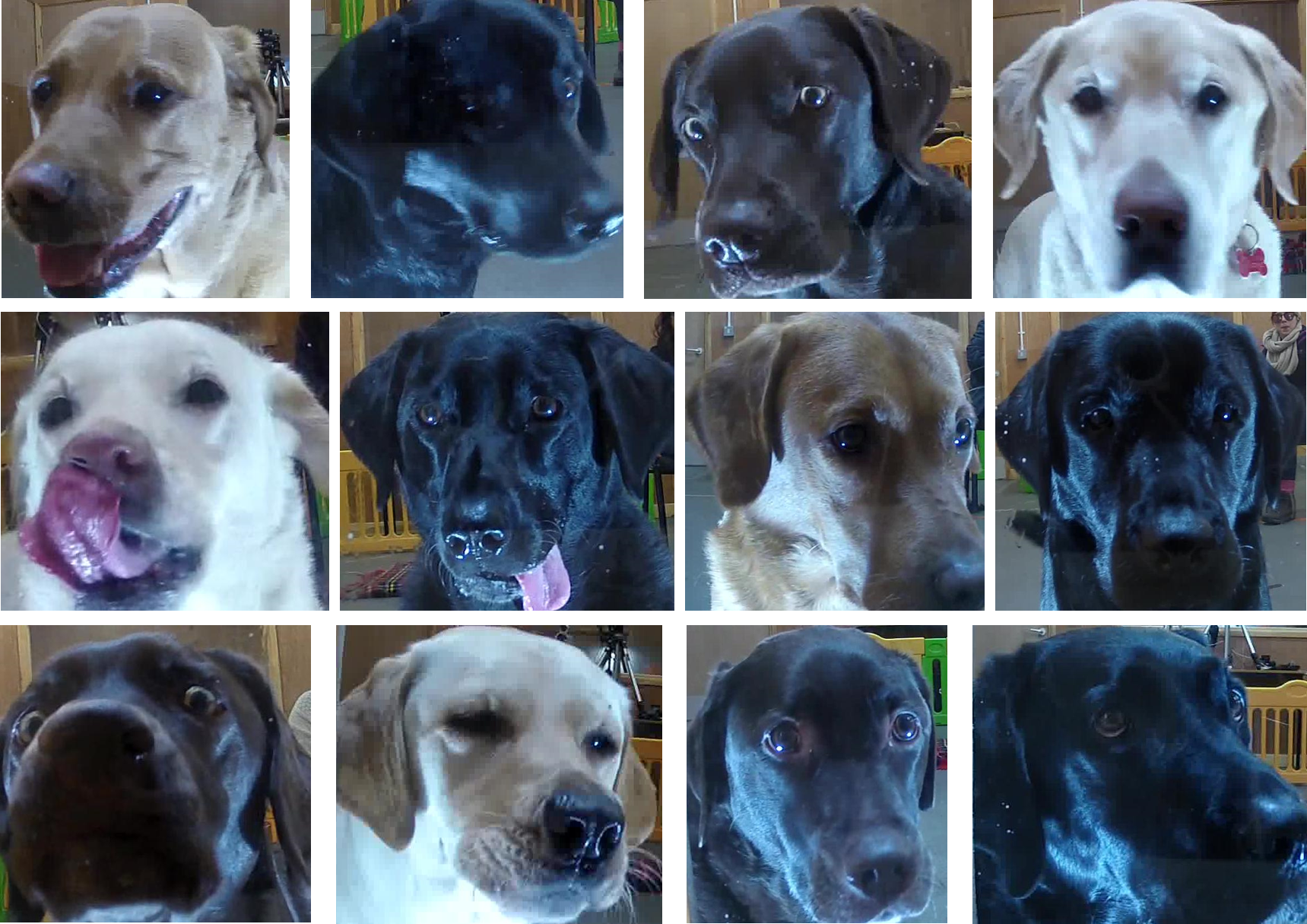}
\caption{\textbf{Example frames from the dataset.} Crops of dog faces extracted from the dataset. The dataset contains images of diverse pose, facial expression, and canine color.}\label{dogs}\vspace{-0.75em}
\end{figure}

\section{Dataset}
{We use the dataset collected by Bremhorst et al.~\cite{bremhorst2019differences}, containing short three-second videos of dogs experiencing the emotional states of frustration and positive anticipation in a controlled experimental setting. They also standardize the dog breed (Labrador Retriever) to reduce the potential effects of morphological variation and extremes on the dogs’ facial expressions.} They used a non-social context in order to eliminate the risk of interference from previously learned attention getting responses: a high-value food reward was used as the triggering stimulus in two conditions – a positive condition predicted to induce positive anticipation, and a negative condition predicted to induce frustration in dogs.

The dataset contains 248 videos of 29 subjects, most of which are neutered. Figure~\ref{age_dog_sex} demonstrates the variety of subjects over age and gender. Each dog was filmed roughly nine times, a third of which in positive anticipation state and two thirds in frustration state. The same emotional state is assumed to hold throughout each video due to their short duration. 
 
The availability of such visual-temp data enables addressing it in two manners: (i) single frames and (ii) \emph{sequences} of frames. The first implies more information loss, but is simpler and more controllable; while the latter includes temporal dimension, which has been shown to have importance for such tasks, e.g., in the context of detection of pain in horses \cite{broome2019dynamics,rashid2022equine}. The prevalent approach in the context of automated recognition of effective states and pain in animals, is, however, the single frame basis (e.g., \cite{hummel2020automatic,andresen2020towards,lu2017estimating,lencioni2021pain}). Due to the exploratory nature of this study, we opt for this option, assuming (at least the majority of) the frames capture information on the emotional state of the dogs. Therefore, our final dataset contains single frames extracted from the videos and labelled accordingly: 12569 negative frames 6823 positive frames. 
Figure~\ref{dogs} presents example faces, indicating the diversity of poses, facial expressions, and subject appearance.

\subsection{Pre-Processing}
The original video frames contain background clutter including the surrounding room, humans, dog body, etc. We aim to focus on the \emph{facial expressions} of the dogs and avoid learning other emotional state predictors (e.g. dog body postures). Hence, we trained a Mask-RCNN~\cite{he17maskrcnn} to identify canine faces, and used it to crop the facial bounding box from each image.
We trained the Mask-RCNN on roughly 200 annotated images from this dataset, making it most suited for this specific experimental setup. Figure~\ref{dogs} shows facial crops acquired using this pre-processing stage.

\section{Experiments and Results}

\subsection{Framework}
We pose the task as a binary classification task distinguishing the two emotional states. We employ the common \emph{"transfer learning"} setup, training a linear probe on top of a \emph{fixed pre-trained backbone} using human annotations. We explore the suitability of different backbones for this task by repeating the experiment with four pre-trained backbones: ResNet and ViT trained either in a supervised manner for image classification~\cite{dosovitskiy2020vit} or in a self-supervised manner using DINO~\cite{caron2021emerging}. 

\subsection{Implementation Details}
The dataset was divided into a training set containing 14830 frames from 22 dog subjects; and a test set containing 4562 frames of 7 dog subjects. Separating the subjects used for training and testing is a common practice in the context of animal face analysis, as it enforces generalization to unseen subjects and ensures that no specific features of an individual are used for classification \cite{andresen2020towards,broome2019dynamics}.

We use ResNet50 architecture for supervised and DINO-trained backbones; \texttt{ViT-S/16} trained in a supervised manner and \texttt{ViT-S/8} trained with DINO. We use  pretrained ViT weights from the Timm Library~\cite{rw2019timm}.
We train all the models for 30 epochs using Adam optimizer~\cite{adam} with betas=($0$, $0.999$) and learning rates: $10^{-4}$ for ResNet backbones and $5\cdot10^{-6}$ for ViT backbone.
We apply several augmentations during training to improve the robustness of the models: horizontal flips, color jitter, and random crops of 80-100\% of the original facial crops. All inputs were resized to size $224\times224$.

\subsection{Results}

\begin{figure}[t!]
\centering
\includegraphics[width=0.5\textwidth]{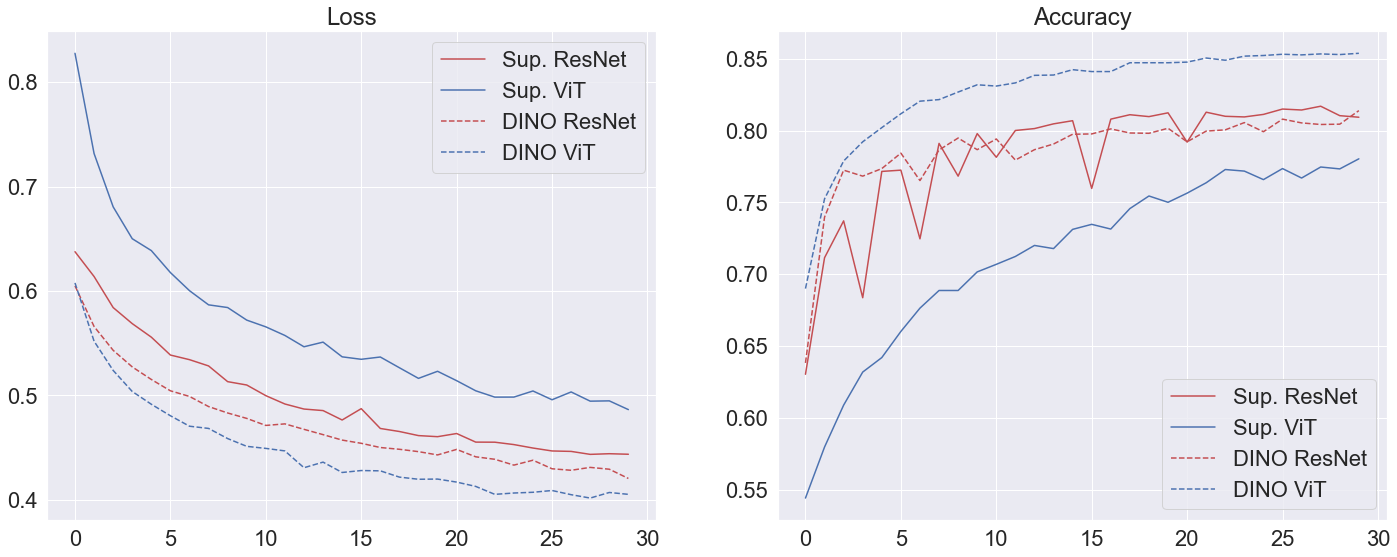}
\caption{\textbf{Loss and Accuracy Curves for each model.} We show the loss and accuracy on the validation set for each trained model. The DINO-ViT based model performs better than models based on other backbones.}\label{loss_acc}\vspace{-0.25em}
\end{figure}

The accuracy measures of our trained models and loss curves appear in Tab.~\ref{table:1} and Fig.~\ref{loss_acc} respectively. The model trained with a DINO-ViT backbone produces the highest accuracy on the validation set. We hypothesize that this is due to DINO-ViT features being sensitive to object parts, as shown in~\cite{amir2021deep}; and due to the nature of the task at hand - emotion classification requires understanding at the object-part level (e.g. states of eyes, ears, etc.). Intriguingly, the backbones pre-trained with DINO produce better results than the supervised backbones.

\begin{table}[t!]
\centering
\begin{tabular}{c| l| c| c} 
 \hline
 \multicolumn{2}{c|}{Backbone} & Train Accuracy & Val. Accuracy \\
 \hline
 \multirow{2}{*}{Sup.} & ResNet50~\cite{he16resnet} & 0.800 & 0.809 \\ 
  \cline{2-4}
  & ViT~\cite{dosovitskiy2020vit} & 0.869 & 0.780  \\
 \hline
 \multirow{2}{*}{DINO}  & ResNet50~\cite{caron2021emerging} & 0.870 & 0.813  \\
  \cline{2-4}
  & ViT~\cite{caron2021emerging} & \textbf{0.878} & \textbf{0.853}  \\
 \hline
\end{tabular}
\caption{\textbf{Classification Results.} The best results are achieved using a pre-trained DINO-ViT as a backbone.}
\label{table:1}
\end{table}

\subsection{Explainability}

\begin{figure*}[t!]
\centering
\includegraphics[width=\textwidth]{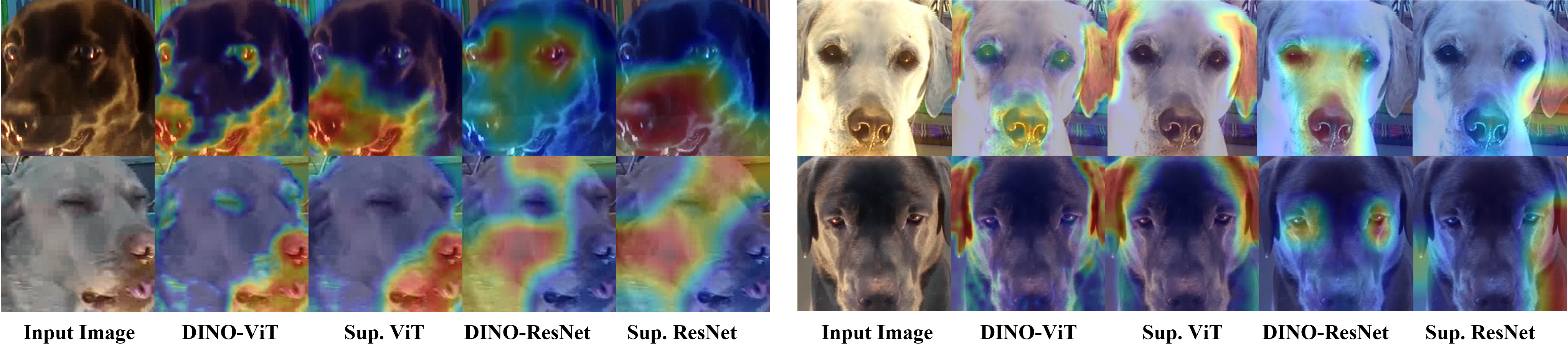}
\caption{\textbf{EigenCAM~\cite{DBLP:journals/corr/abs-1910-01279} activation maps.} on several images for our four different models. The images in the top and bottom row have positive and negative ground-truth emotions respectively. The DINO-ViT backbone addresses similar areas to those proposed by human annotated action units.}\label{dog_interp}\vspace{-0.25em}
\end{figure*}

We further investigate our trained models by visualizing their activation maps on several images. We apply Eigen-CAM~\cite{DBLP:journals/corr/abs-1910-01279} to visualize the principal components of the final activations for each model.
It has been shown that Eigen-CAM provides more easily interpretable results with less computation compared to other CAM methods such as the popular Grad-CAM \cite{SelvarajuCDVPB17}. We chose this method since unlike other visualization methods such as Grad-CAM \cite{DBLP:journals/corr/abs-1910-01279} and Grad-CAM++ \cite{8354201}, Eigen-CAM is a class-independent tool. This property enables Eigen-CAM to  visualize learned patterns even when the model prediction is wrong, as opposed to older CAM methods that produce irrelevant maps when their prediction isn't correct. This property of Eigen-CAM enables interpreting reasons for prediction failure. It is more consistent and class discriminative compared to other state of the art visualization methods.In addition, EigenCAM is not model-specific - it can be used for both ViTs and CNNs without changing layers.

Several qualitative examples are presented in Fig.~\ref{dog_interp}. We observe several characteristics common to all activation maps of each pretrained model - (i) The ViT models seem to exhibit better localization than the ResNet models. The highly activated regions (marked by red) are smaller and lay on more salient regions (e.g. eyes, ears, nose rather than skin). (ii) The DINO-ViT model seems to activate on multiple salient regions rather than one (e.g. activating on ears, eyes and nose rather than just ears on the top-right example). We attribute the success of ViT based models to the ability of ViTs to provide a more localized signal than the ResNet models. This stems in their architecture -- the resolution of ViT features remains constant throughout the layers, while the resolution of CNN features diminishes as the layers become deeper. 

\section{Discussion}
Despite the huge (and increasing) body of work on emotions in animals, there is still no common agreement between researchers even on the most basic questions. For instance, already Darwin suggested that some emotional cues (such as facial expressions) may have visual similarity across different species, and even bear the same meaning '\cite{Darwin1872}; however, recent research applying objective tools (such as AnimalFACS) for measuring facial expressions has begun to question  this assumption~\cite{caeiro2017development}.  

On a practical level, objective assessment of animal emotion should be 
a cornerstone of animal welfare practices, focusing not only on reduction of negative emotions, but also recently attempting to promote positive states. The subtle and complex nature of animal emotions, and at least some of their characteristics being species-specific (and thus different from humans') pose significant challenges.

Deep learning has the potential to be a game changer in providing answers both to foundational scientific questions on the nature of animal emotions, as well as pushing forward practical tools in the context of animal welfare and health. We have explored here in detail how it can be used to classify two specific emotional states in dogs of positive and negative valence. More specifically, having examined the suitability of different pre-trained backbones for this task, we conclude that DINO-ViT features have superior performance in this context, possibly due to the DINO-ViT features being sensitive to object parts~\cite{amir2021deep}

However, for deep learning models to contribute both to scientific discoveries in animal emotion, and to applied tools for animal welfare and healthcare, these models judgment should be \emph{interpretable}. We show existing explainability methods hold much promise for this task. 
One interesting possibility is to  include mapping features learnt by the deep learning methods to concepts grounded in behavioral meaning such as DogFACS \cite{caeiro2017dogs} action units, in a way similar to what was done for human FACS in \cite{khorrami2015deep}.
In any case, the results presented will serve as first baseline for future research into canine affective computing using deep learning techniques. 

\section*{Acknowledgements}
The authors would like to thank Prof. Hanno W\"urbel for his guidance in collecting and analyzing the data used in this study. 

The research was partially supported by the grant from the Ministry of Science and Technology of Israel and RFBR according to the research project no. 19-57-06007, and by the Israel Ministry of Agriculture and Rural Development.

{\small
\bibliographystyle{ieee_fullname}
\bibliography{egbib}
}

\end{document}